\pdfoutput=1

\documentclass[11pt]{article}

\usepackage{ACL2023}

\usepackage{times}
\usepackage{latexsym}

\usepackage{graphicx}
\usepackage{subcaption,booktabs}
\usepackage{tabularx}
\usepackage{multirow, amsmath}
\usepackage{placeins}
\usepackage{gensymb} 
\usepackage{amssymb}

\usepackage{amsthm}
\usepackage{color, colortbl}
\usepackage{float}
\usepackage{enumitem}

\usepackage[T1]{fontenc}

\usepackage[utf8]{inputenc}

\usepackage{microtype}

\usepackage{inconsolata}

\title{The Magic of \texttt{IF}: 
Investigating Causal Reasoning Abilities in Large Language Models of Code}

\author{
	Xiao Liu$^{1}$, 
        Da Yin$^{2}$,
        Chen Zhang$^{1}$,
	Yansong Feng$^{1,3}$\thanks{\;\;Corresponding author.}\and
	Dongyan Zhao$^{1,4,5}$ \\
        $^1$Wangxuan Institute of Computer Technology, Peking University\\
	$^2$Computer Science Department, University of California, Los Angeles\\
	$^3$The MOE Key Laboratory of Computational Linguistics, Peking University\\
	$^4$Beijing Institute for General Artificial Intelligence\\
	$^5$State Key Laboratory of Media Convergence Production Technology and Systems\\
	{\tt \{lxlisa,zhangch,fengyansong,zhaody\}@pku.edu.cn} \\
	{\tt da.yin@cs.ucla.edu}\\
}

\begin{document}
\maketitle

\begin{abstract}
Causal reasoning, the ability to identify cause-and-effect relationship, is crucial in human thinking. Although large language models (LLMs) succeed in many NLP tasks, it is still challenging for them to conduct complex causal reasoning like abductive reasoning and counterfactual reasoning. Given the fact that programming code may express causal relations more often and explicitly with conditional statements like \textbf{\texttt{if}}, we want to explore whether Code-LLMs acquire better causal reasoning abilities. Our experiments show that compared to text-only LLMs, Code-LLMs with code prompts are significantly better in causal reasoning. 
We further intervene on the prompts from different aspects, and discover that the programming structure is crucial in code prompt design, while Code-LLMs are robust towards format perturbations.
Code and data are available at \url{https://github.com/xxxiaol/magic-if}.
\end{abstract}
\section{Introduction}
Human beings rely heavily on the capacity for \emph{causal reasoning}~\cite{sloman2005causal,hagmayer2007causal}. People understand the observed facts, predict future events, and speculate about what might have happened if things had been different with the help of their causal reasoning skills. For instance, when we go home and find a mess, we probably want to figure out why it happened. If we determine that a bird flew into the house, we might then consider whether the mess could have been avoided if we had closed the window.

\begin{figure}[t]
    \centering
    \includegraphics[width=\columnwidth]{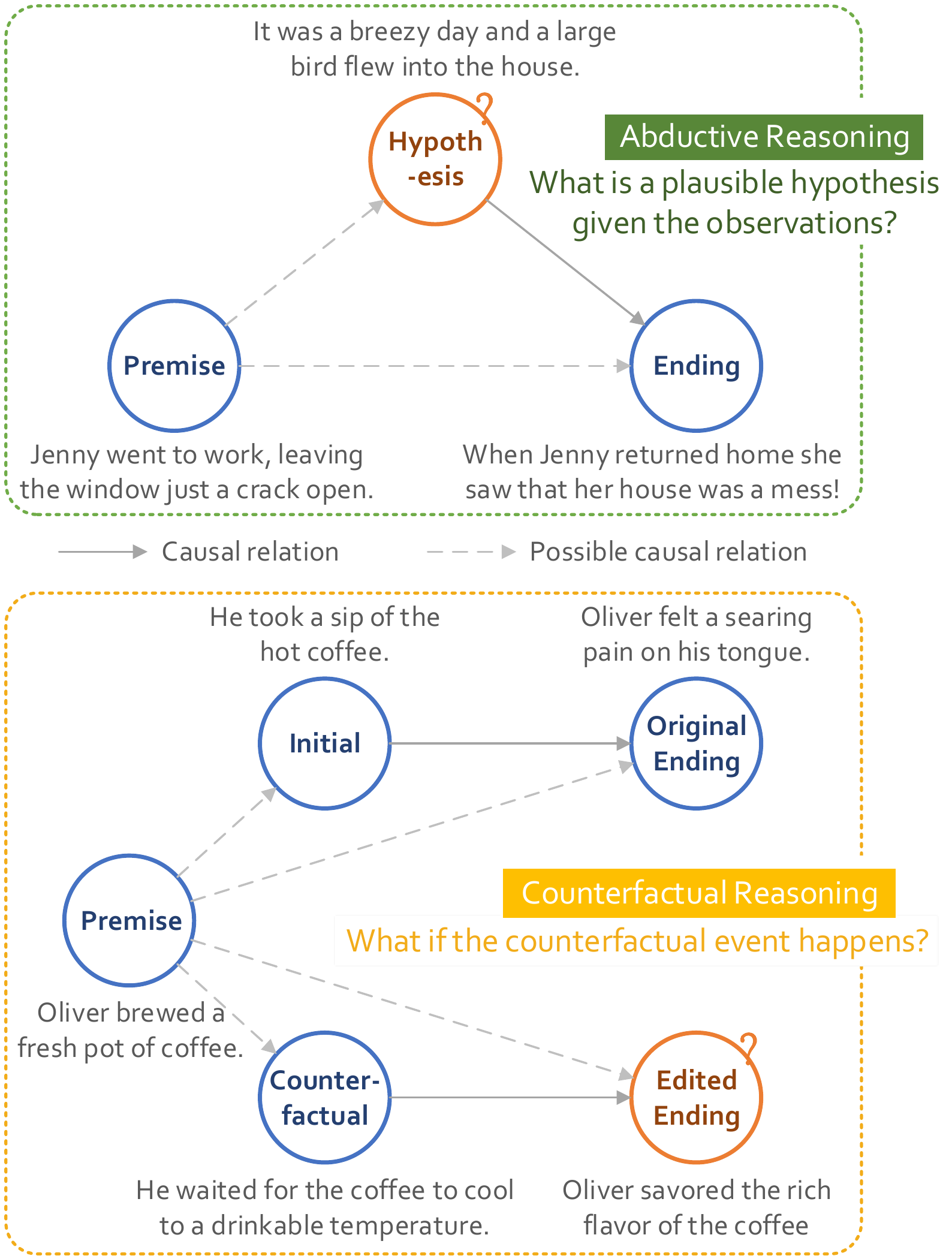}
    \caption{Causal relationships between events in two causal reasoning tasks.}
    \label{fig-intro}
    \vspace{-1mm}
\end{figure}

\begin{figure*}[th]
    \centering
    \includegraphics[width=\textwidth]{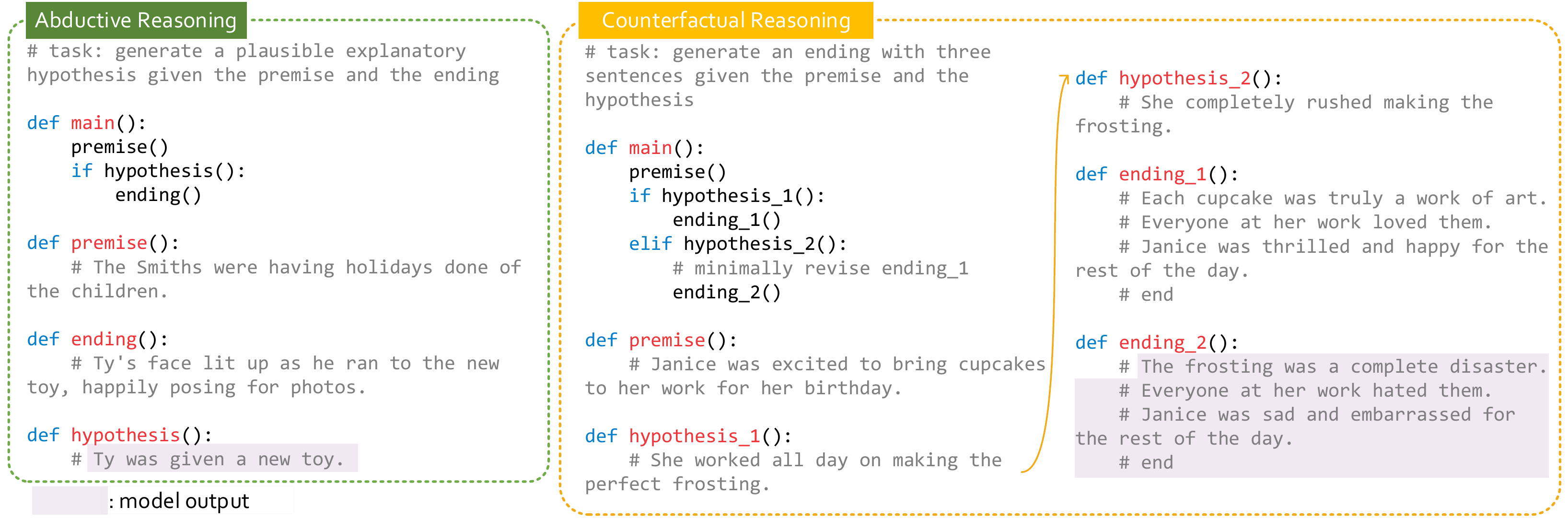}
    \caption{Example code prompts of abductive reasoning and counterfactual reasoning.} 
    \label{fig-prompt}
    \vspace{-1mm}
\end{figure*}

Although large language models (LLMs) demonstrate great language understanding and generation abilities, it is still challenging for them to perform complex causal reasoning such as the example above. Powerful LLMs are able to understand single cause-and-effect relations~\cite{brown2020language, wang2021towards}, like \emph{a man losing his balance} causes him to \emph{fell}. However, when it comes to more complex causal structures involving multiple events and alternative branches (like \emph{close the window} or not), 
LLMs perform much inferior to humans~\cite{bhagavatula2019abductive, qin2019counterfactual}.
In this paper, we consider two challenging causal reasoning tasks: abductive reasoning and counterfactual reasoning. 
Abductive reasoning requires models to generate a plausible reason for the \emph{ending} while being consistent with the \emph{premise}. 
Counterfactual reasoning asks what will occur in the \emph{counterfactual branch}. Causal relationships between events in these tasks are shown in Figure~\ref{fig-intro}.

A potential difficulty for LLMs to learn complex causal structures is that they are rarely expressed explicitly in the text. News articles or narratives may contain multiple events with causal relationships, like an incident and a chain of consequences.
However, these events are often written chronologically, and {it is hard to extract the causal structure from the text without further annotation}.
Branches are expressed rarer in text, except for the multi-branching storytelling style~\cite{nisi2006weird}.

On the other hand, causal relations are exhibited more commonly in code. Conditional statements like \texttt{if} direct the computer to execute certain commands, provided a condition is met. 
This explicitly demonstrates the causal relationship between the \emph{condition block} and the \emph{execution block}. 
Code can also express branching with \texttt{elif} or \texttt{switch} statements, and the nesting feature enables code to describe more complex structures\footnote{Although causal expressions like \emph{if} are also used in natural languages, representing complex causal structures in text is not as clear and structured as in code.}.

This motivates us to utilize code models in natural language causal reasoning.
Recently, large language models of code (Code-LLMs) are receiving increasing attention~\cite{chen2021evaluating,xu2022systematic}. They exhibit strong code generation performance, and their structural prediction abilities help complete structural natural language tasks like argument graph generation~\cite{madaan2022language} and event argument extraction~\cite{wang2022code4struct}.
Being pre-trained on code with abundant causal expressions, Code-LLMs may also have gained better causal reasoning abilities. 

We conduct experiments on the unsupervised abductive reasoning and counterfactual reasoning tasks. To generate task outputs, we design code prompts like Figure~\ref{fig-prompt} to clearly represent the causal structures of the tasks. Results show that Code-LLMs with code prompts perform much better than text-only LLMs and previous methods.
To better understand why the code prompts are effective,
we break down the prompts and analyze the influence of different aspects. We find that Code-LLMs are very sensitive to the \emph{programming structure} (specifically, the conditional statements), while being robust towards format perturbations and programming language changes.

Our main contributions are as follows:
1) We design code prompts to tackle causal reasoning tasks, by leveraging conditional statements in code to represent causal structures. 
2) We evaluate Code-LLMs with code prompts on the abductive reasoning and counterfactual reasoning tasks, and exhibit that code models with code prompts are better causal reasoners than text models.
3) We break down the code prompt in detail and find that the programming structure is crucial to the performance.
\section{Modeling Causal Structure with Code}
\begin{table*}[!ht]
    \centering
    \small
    \scalebox{0.9}{
    \begin{subtable}[t]{0.57\textwidth}
        \centering
        \setlength{\tabcolsep}{3.5pt}
        \begin{tabular}{lcccc}
        \toprule
        & \textbf{BLEU$_4$} & \textbf{ROUGE$_\textrm{L}$} & \textbf{CIDEr} & \textbf{BERTScore} \\
        \midrule
        \textsc{Delorean} & 1.6 & 19.1 & 7.9 & 41.7 \\
        \textsc{Cold} & 1.8 & 19.5 & 10.7 & 42.7 \\
        \textsc{Diffusion} & 7.1 & 28.3 & 30.7 & \quad-\:\:\: \\
        \midrule
        \textsc{Davinci$_{002}$} & 4.9 & 27.0 & 26.6 & 56.8 \\
        \textsc{Davinci$_{003}$} & 4.6 & 25.8 & 10.7 & 57.1 \\
        \textsc{Codex} & \textbf{13.7} & \textbf{39.6} & \textbf{81.8} & \textbf{64.9} \\
        \bottomrule
        \end{tabular}
        \caption{Abductive reasoning.}
        \label{table-abductive}
    \end{subtable}
    \hspace{0.5em}
    \begin{subtable}[t]{0.41\textwidth}
        \centering
        \setlength{\tabcolsep}{3.1pt}
        \begin{tabular}{lccc}
        \toprule
        & \textbf{BLEU$_4$} & \textbf{ROUGE$_\textrm{L}$} & \textbf{BERTScore} \\
        \midrule
        \textsc{Delorean} & 21.4 & 40.7 & 63.4 \\
        \textsc{CGMH} & 41.3 & - & 73.8 \\
        \textsc{EduCat} & 44.1 & - & 74.1 \\
        \midrule
        \textsc{Davinci$_{002}$} & 49.0 & 54.7 & 73.0 \\
        \textsc{Davinci$_{003}$} & 30.6 & 45.2 & 69.4 \\
        \textsc{Codex} & \textbf{66.8} & \textbf{70.0} & \textbf{82.5} \\
        \bottomrule
        \end{tabular}
        \caption{Counterfactual reasoning.}
        \label{table-counterfactual}
    \end{subtable}
    }
    \caption{Automatic evaluation results on two unsupervised causal reasoning tasks in the zero-shot setting. Numbers are in percentages (\%).}
    \label{table-main}
\end{table*}

\begin{table*}[ht]
    \centering
    \small
    \scalebox{0.9}{
    \begin{tabular}{lccc}
    \toprule
    & \textbf{\textsc{Codex}} & \textbf{Neutral} & \textbf{\textsc{Davinci$_{002}$}} \\
    \midrule
    \rowcolor[gray]{0.95} \textbf{Abductive Reasoning} & & & \\
    Coherence with Premise & \textbf{34\%} & 48.5\% & 17.5\%\\
    Coherence with Ending & \textbf{32\%} & 42.5\% & 25.5\%\\
    Overall Coherence & \textbf{40\%} & 38\% & 22\%\\
    \rowcolor[gray]{0.95} \textbf{Counterfactual Reasoning} & & & \\
    Coherence & \textbf{36.5\%} & 39.5\% & 24\%\\
    Preservation & \textbf{47.5\%} & 39.5\% & 13\%\\
    \bottomrule
    \end{tabular}
    }
    \caption{Human evaluation of comparing \textsc{Codex} and \textsc{Davinci$_{002}$}.}
    \label{table-human}
\end{table*}

We convert the input of causal reasoning tasks into the form of code prompt for Code-LLMs to understand better.
We expect the prompts to meet two requirements: 1) clearly represent the causal relationships between events, and 2) as most Code-LLMs only support generating at the end, the target output should appear at the end of the prompts.
The first requirement is addressed with conditional statements. However, for the second, the target prediction is not always the last part of the conditional statements, e.g., in abductive reasoning we want to predict the hypothesis, which is the condition in the \texttt{if} structure. To address this, we uniformly use functions to represent events. 
As shown in Figure~\ref{fig-prompt}, the causal structure is described in the \texttt{main} function.
All the event functions are listed afterwards, leaving the target event function at the last.

\vspace{1mm}
\noindent\textbf{Abductive Reasoning.}
Abductive reasoning requires models to generate a plausible hypothesis $H$ given the observations: premise $P$ and ending $E$. The chronological order of these three events is $P \rightarrow H \rightarrow E$, and the hypothesis causes the ending to occur.

In Figure~\ref{fig-prompt}, we regard the task definition as an instruction and place it as a comment at the beginning of the prompt. The causal structure is represented in the \texttt{main} function like: executing the premise, and if the hypothesis is met, executing the ending\footnote{Although not entirely accurate, this approximates the actual underlying causal relationships.}. The content of each event is presented as a comment of its function. The \texttt{hypothesis} function is placed at the last, leaving for models to complete. The generation process stops with a line break.

\vspace{1mm}
\noindent\textbf{Counterfactual Reasoning.}
Counterfactual reasoning aims to rewrite a story under a counterfactual condition. As in Figure~\ref{fig-intro}, the input consists of four parts: the premise $P$, the initial context $C_1$, the original ending $E_1$, and the counterfactual context $C_2$. Models are asked to generate the counterfactual ending $E_2$ that \emph{minimally} modifies the original ending $E_1$  and is coherent with the counterfactual context $C_2$. 

The causal relationships are represented with the \texttt{if-elif} structure. The premise $P$ is executed first, and then if the initial context  $C_1$ is met, the original ending $E_1$ is executed; otherwise, if the counterfactual context $C_2$ is met, the counterfactual ending $E_2$ will be executed. For ease of exposition, we call the context \texttt{hypothesis} as well, being consistent with the former task. The event contents are also written as comments for event functions. We use \texttt{\# end} to mark the finish of the ending.

\section{Evaluation}
\begin{table*}[!ht]
    \centering
    \small
    \scalebox{0.9}{
    \begin{subtable}[t]{0.57\textwidth}
        \centering
        \setlength{\tabcolsep}{3pt}
        \begin{tabular}{lcccc}
        \toprule
        & \textbf{BLEU$_4$} & \textbf{ROUGE$_\textrm{L}$} & \textbf{CIDEr} & \textbf{BERTScore} \\
        \midrule
        \textsc{Codex$_{text}$} & 11.7 & 37.5 & 78.5 & 62.5 \\
        \textsc{Codex$_{code}$} & 13.7 & 39.6 & 81.8 & 64.9 \\
        \textsc{Codex$^*_{code}$} & \textbf{16.5} & \textbf{42.0} & \textbf{91.6} & \textbf{66.3} \\
        \midrule
        \textsc{Davinci$_{text}$} & 4.9 & 27.0 & 26.6 & 56.8 \\
        \textsc{Davinci$_{code}$} & 6.7 & 31.1 & 46.2 & 59.9 \\
        \textsc{Davinci$^*_{code}$} & \textbf{9.0} & \textbf{35.0} & \textbf{64.0} & \textbf{62.2} \\
        \bottomrule
        \end{tabular}
        \caption{Abductive reasoning.}
        \label{table-exchange-abductive}
    \end{subtable}
    \hspace{0.5em}
    \begin{subtable}[t]{0.41 \textwidth}
        \centering
        \setlength{\tabcolsep}{3.1pt}
        \begin{tabular}{lccc}
        \toprule
        & \textbf{BLEU$_4$} & \textbf{ROUGE$_\textrm{L}$} & \textbf{BERTScore} \\
        \midrule
        \textsc{Codex$_{text}$} & 55.1 & 61.3 & 77.8 \\
        \textsc{Codex$_{code}$} & 66.8 & 70.0 & 82.5 \\
        \textsc{Codex$^*_{code}$} & \textbf{73.3} & \textbf{74.7} & \textbf{85.3} \\
        \midrule
        \textsc{Davinci$_{text}$} & \textbf{49.0} & \textbf{54.7} & \textbf{73.0} \\
        \textsc{Davinci$_{code}$} & 40.4 & 48.5 & 70.5 \\
        \textsc{Davinci$^*_{code}$} & 43.7 & 52.0 & 72.8 \\
        \bottomrule
        \end{tabular}
        \caption{Counterfactual reasoning.}
        \label{table-exchange-counterfactual}
    \end{subtable}
    }
    \caption{Effect of exchanging prompts for \textsc{Codex} and \textsc{Davinci$_{002}$} (\%). $^*$ indicates the best code prompt experimented in Section~\ref{sec-aspect}.}
    \label{table-exchange}
\end{table*}
\noindent\textbf{Datasets.}
We experiment on the {\usefont{T1}{pzc}{m}{n} ART} dataset~\cite{bhagavatula2019abductive} for the evaluation of abductive reasoning, and the TimeTravel dataset~\cite{qin2019counterfactual} for counterfactual reasoning, with 3,561 and 1,871 test instances, respectively.

\noindent\textbf{Models.} 
We experiment with \textsc{Codex}~\cite{chen2021evaluating}, trained on a blend of code and text, as the Code-LLM. The specific version is \texttt{code-davinci-002}. 
We compare with two LLMs: the latest versions of GPT-3~\cite{brown2020language} \texttt{text-davinci-002} and \texttt{text-davinci-003} (referred to as \textsc{Davinci$_{002}$} and \textsc{Davinci$_{003}$}). Both of them originate from \textsc{Codex} and are tuned with instructions. We follow OpenAI's default settings in \textsc{Codex} and \textsc{Davinci} decoding, and the text prompts for \textsc{Davinci} are in Figure~\ref{fig-text}.

We also compare with previous unsupervised methods on these tasks, including \textsc{Delorean}~\cite{qin2020back}, \textsc{Cold}~\cite{qin2022cold}, \textsc{Diffusion}~\cite{li2022diffusion}, CGMH~\cite{miao2019cgmh}, and \textsc{EduCat}~\cite{chen2022unsupervised}\footnote{All these methods except \textsc{Diffusion} use GPT-2~\cite{radford2019language} as the base model, and the model size ranges from medium to XL.}. Appendix~\ref{sec-appendix-model} provides a brief introduction of these methods. 

\vspace{1mm}
\noindent\textbf{Automatic Evaluation.}
We use the following automatic evaluation metrics:
BLEU$_4$~\cite{papineni2002bleu}, ROUGE$_\textrm{L}$~\cite{lin2004rouge}, CIDEr~\cite{vedantam2015cider} and BERTScore~\cite{zhang2019bertscore} based on BERT-base for abductive reasoning; BLEU$_4$, ROUGE$_\textrm{L}$ and BERTScore for counterfactual reasoning.

Table~\ref{table-main} reports the automatic evaluation results in the zero-shot setting. \textsc{Codex} significantly
outperforms previous methods and \textsc{Davinci} on both tasks (with significance level $\alpha = 0.01$), exhibiting strong causal reasoning ability.
Although the two \textsc{Davinci} models are based on \textsc{Codex}, their causal reasoning abilities may be weakened during instruction tuning, and this phenomenon is called \emph{alignment tax}~\cite{ouyang2022training}.
\textsc{Davinci$_{003}$} underperforms \textsc{Davinci$_{002}$} on most metrics, probably because it tends to generate longer and more discursive outputs, which do not comply with the tasks.

\begin{table*}[th]
    \centering
    \small
    \scalebox{0.9}{
    \begin{tabular}{llcccc}
    \toprule
    & & \textbf{BLEU$_4$} & \textbf{ROUGE$_\textrm{L}$} & \textbf{CIDEr} & \textbf{BERTScore} \\
    \midrule
    & \textsc{Codex} & 13.7 & 39.6 & 81.8 & 64.9 \\
    \midrule 
    \rowcolor[gray]{0.95} &  No Instruction & 12.1 & 37.4 & 73.8 & 62.9 \\
    \rowcolor[gray]{0.95} \multirow{-2}{*}{\textbf{Information}} & Function Name Perturbation & 15.1 & 39.1 & 77.8 & 64.6 \\
    & Sequential Structure & 9.6 & 36.8 & 72.0 & 63.5 \\
    \multirow{-2}{*}{\textbf{Structure}} & Disruption & 7.9 & 30.3 & 49.8 & 58.5 \\
    \rowcolor[gray]{0.95} & Class & 16.0 & 41.0 & 87.4 & 65.8 \\
    \rowcolor[gray]{0.95}  & Print & 13.8 & 39.4 & 82.0 & 65.0 \\
    \rowcolor[gray]{0.95} \multirow{-3}{*}{\textbf{Format}} & Return & 13.0 & 40.3 & 83.4 & 65.5 \\
    & Java & \textbf{16.5} & \textbf{42.0} & \textbf{91.6} & \textbf{66.3} \\
    \multirow{-2}{*}{\textbf{Language}} & C & 15.5 & 41.0 & 88.0 & 65.6 \\
    \bottomrule
    \end{tabular}
    }
    \caption{Intervention results on abductive reasoning (\%).}
    \label{table-aspect-a}
\end{table*}
\vspace{1mm}
\noindent\textbf{Human Evaluation.}
We conduct pairwise comparison between \textsc{Codex} and \textsc{Davinci$_{002}$} on 100 test examples. Annotators are asked to choose the better output given the task requirements. For abductive reasoning, the outputs are rated from three aspects: coherence with the premise, coherence with the ending, and the overall coherence. For counterfactual reasoning, the outputs are rated from coherence with the context and the extent of preserving the original ending. Each example is rated by at least two annotators, and the average inter-rater reliability is 0.64.

The results are shown in Table~\ref{table-human}. \textsc{Codex} outperforms \textsc{Davinci$_{002}$} in all aspects. It better considers the context in generation, and is able to preserve the original content in counterfactual reasoning.

\vspace{1mm}
\noindent\textbf{Contributions of the Model and the Prompt.}
We exchange the prompts of code and text models, to measure the contributions of the model and the prompt. The results are in Table~\ref{table-exchange}. We find that \textsc{Codex} performs better with the code prompt, as the code prompt clearly describes the causal relation between events. Code prompts benefit the text model \textsc{Davinci$_{002}$} on abductive reasoning, but have negative impacts on counterfactual reasoning. A possible reason is that the causal structure in counterfactual reasoning is more complicated, leading to a more complex code which is harder for text models to understand. 
\section{What are Crucial in Code Prompts?}
\label{sec-aspect}

To paint a better picture of the key points in the code prompts, we intervene on the prompts from four aspects and measure the influences of the interventions.
The four aspects we select are \emph{information}, \emph{structure}, \emph{format}, and \emph{language}. The former two, the prior information provided and the programming structure of functions, are content-related; the latter two, the code format and programming languages, are form-related. An ideal model should rely on the content and be insensitive to form perturbations. The interventions are described below, with examples in Figure~\ref{fig-aspect}.

\vspace{1mm}
\noindent\textbf{\textit{Information.}} We study two types of prior information: task instructions and function names. In \emph{No Instruction}, we remove the task instruction from the prompts. In \emph{Function Name Perturbation}, we replace original function names with anonymous \texttt{functionX}. For example, we replace \texttt{premise()} and \texttt{hypothesis()} in Figure~\ref{fig-prompt} with \texttt{functionA()} and \texttt{functionB()}, respectively. It eliminates the information in function names and only allows models to learn the event relations from programming structures.

\vspace{1mm}
\noindent\textbf{\textit{Structure.}} The first way to intervene in the programming structure is to convert the conditional structures into  sequential structures, referred to as \emph{Sequential Structure}. The events are executed sequentially, like \texttt{premise(), hypothesis(), ending()} in abductive reasoning. In the second way called \emph{Disruption}, we randomly disrupt the positions of the functions in the conditional structure. 
For instance, \texttt{if hypothesis(): ending()} can be disrupted into \texttt{if ending(): hypothesis()}. We also apply the function name perturbation in disruption to eliminate the impact of function names.

\vspace{1mm}
\noindent\textbf{\textit{Format.}} We test three formats besides the original one: \emph{Class}, \emph{Print} and \emph{Return}. The first one converts the original code into a class.
We define the programming structure in the \texttt{\_\_init\_\_} method, and move the event functions into the class. In \emph{Print}, we represent the content of events as a string and print it in the function body, like \texttt{def premise(): print("The Smiths ...")}. And in \emph{Return}, the string is the return value of event functions.

\vspace{1mm}
\noindent\textbf{\textit{Language.}} We also convert the original Python programs into two other languages, \emph{Java} and \emph{C}, to evaluate the influence of programming languages.

\vspace{1.5mm}
\noindent\textbf{Intervention Results.}
We evaluate the influence of interventions on abductive reasoning in Table~\ref{table-aspect-a}, and the results on counterfactual reasoning are in Table~\ref{table-aspect-c}. 
The absence of prior information causes a small decrease in results. Even if the instruction or function names are not provided, \textsc{Codex} is able to perform causal reasoning based on conditional statements.
Changes in the programming structure have a larger negative impact. Comparing \emph{Function Name Perturbation} and \emph{Disruption}, the alteration of two characters (swap \texttt{B} and \texttt{C} in \texttt{functionB} and \texttt{functionC}) results in a major drop, showing that the conditional structure that reasonably depicts the relations between events is crucial in \textsc{Codex} reasoning.

\textsc{Codex} is quite robust towards format and language changes. Settings like \emph{Class} and \emph{Java} are even better than the original one, revealing that the performance can be further improved with delicate prompt engineering.
\section{Conclusion}
We investigate the causal reasoning ability of Code-LLMs. With code prompts of conditional statements, Code-LLMs achieve great performance in abductive and counterfactual reasoning, outperforming text-only LLMs significantly. Our study on different aspects of code prompts shows that providing a reasonable causal structure in code can help generate plausible outputs, and Code-LLMs are robust towards format perturbations.

\section*{Limitations}
\paragraph{Language} Our experiments are conducted on English, as all Code-LLMs we know are pre-trained on English programming languages. Fundamentally, most popular programming languages are English-based, but international programming languages (which work in multiple languages) like Scratch, or non-English-based programming languages like Qalb also emerge. We look forward to the appearance of Code-LLMs on these programming languages.

\paragraph{Prompt Engineering} We manually design the prompts without prompt engineering techniques such as prompt search. The searched prompts may outperform the ones we used, but our experiments on interventions show that \textsc{Codex} is fairly robust towards format perturbations.

\paragraph{Model}
\begin{table*}[!ht]
    \centering
    \small
    \scalebox{0.9}{
    \begin{subtable}[t]{0.57\textwidth}
        \centering
        \setlength{\tabcolsep}{3.5pt}
        \begin{tabular}{lcccc}
        \toprule
        & \textbf{BLEU$_4$} & \textbf{ROUGE$_\textrm{L}$} & \textbf{CIDEr} & \textbf{BERTScore} \\
        \midrule
        \textsc{Codex} & \textbf{15.0} & \textbf{39.8} & \textbf{82.2} & \textbf{67.8} \\
        \textsc{ChatGPT} & 5.1 & 26.9 & 17.5 & 62.6 \\
        \textsc{GPT-4} & 6.3 & 29.2 & 27.8 & 65.1 \\
        \textsc{Bard} & 5.7 & 31.5 & 14.8 & 66.0 \\
        \bottomrule
        \end{tabular}
        \caption{Abductive reasoning.}
    \end{subtable}
    \hspace{0.5em}
    \begin{subtable}[t]{0.41\textwidth}
        \centering
        \setlength{\tabcolsep}{3.1pt}
        \begin{tabular}{lccc}
        \toprule
        & \textbf{BLEU$_4$} & \textbf{ROUGE$_\textrm{L}$} & \textbf{BERTScore} \\
        \midrule
        \textsc{Codex} & \textbf{68.4} & \textbf{70.3} & \textbf{84.7} \\
        \textsc{ChatGPT} & 15.3 & 34.7 & 70.0 \\
        \textsc{GPT-4} & 38.5 & 55.5 & 78.6 \\
        \textsc{Bard} & 12.1 & 22.0 & 62.1 \\
        \bottomrule
        \end{tabular}
        \caption{Counterfactual reasoning.}
    \end{subtable}
    }
    \caption{Automatic evaluation results on a subset of 100 instances in the zero-shot setting. Numbers are in percentages (\%).}
    \label{table-extra}
\end{table*}

LLMs update quickly. From the time we submitted the paper until now, several new LLMs have been released. We try to compare their performance with ours. We select three new LLMs: \textsc{ChatGPT}, \textsc{GPT-4}~\cite{openai2023gpt}, and \textsc{Bard}\footnote{Experiments are done with models updated to May 10, 2023.}, and feed the text prompts to them. Because we do not have access to some of their APIs, we only experiment on a subset of 100 instances and report the results in Table~\ref{table-extra}. \textsc{Codex} outperforms all these models in the automatic evaluation, but part of the reason is that these models provide more detailed outputs than the reference. We provide a case study in Appendix~\ref{sec-appendix-case}.

Since \textsc{Codex} is no longer available to the public, we provide \textsc{Codex} generation results in our GitHub repository. We also looked for alternatives and tried two open source Code-LLMs \textsc{CodeGen}~\cite{nijkamp2022codegen} (version CodeGen-16B-Mono) and \textsc{StarCoder}~\cite{li2023starcoder} with our code prompts. However, as shown in the case study, their performance is not comparable to \textsc{Codex}, probably because they are more than ten times smaller in size.
\section*{Ethics Statement}
Our work is based on off-the-shelf LLMs. As the results may inherit the underlying bias of LLMs, they cannot be used individually without human supervision. 
The Codex API was free when the experiments were conducted, and the Davinci APIs cost \$0.02 per thousand tokens. We conduct all the experiments with less than \$100.
We recruit annotators for human evaluation from friends and colleagues of authors. All annotators are fairly paid with more than \$10 per hour. 

\section*{Acknowledgments}
This work is supported in part by NSFC (62161160339). We would like to thank the anonymous reviewers for the helpful discussions and suggestions. For any correspondence, please contact Yansong Feng.

\bibliography{custom}
\bibliographystyle{acl_natbib}

\clearpage
\appendix
\section{Appendix}
\setcounter{figure}{0}                     
\renewcommand\thefigure{A.\arabic{figure}}
\setcounter{table}{0}                     
\renewcommand\thetable{A.\arabic{table}}
\label{sec-appendix}
\subsection{Related Work}
\paragraph{Causal Reasoning}
There is a growing interest in the NLP community to equip models with causal reasoning abilities.~\citet{chang2005causal,gordon2011commonsense} measure causality between words and phrases with statistical methods,~\citet{rink2010learning, li2019knowledge} use explicit semantic cues, and~\citet{liu2021everything, zhang2022rock} discover causal relations with causal inference methods like propensity score matching.~\citet{li2019learning} finetune LLMs on causal event corpus, and~\citet{du2021excar,wang2022care} augment LLMs with causal knowledge graphs.
Contrast to them, we explore the causal reasoning abilities acquired by Code-LLMs in pre-training.

\paragraph{Applying Code-LLMs to Natural Language Tasks}
With the recent development of Code-LLMs, several works attempt to solve natural language tasks with code models. They mainly focus on two areas: numerical reasoning and structural prediction.~\citet{gao2022pal, chen2022program, wu2022autoformalization} apply Code-LLMs to numerical reasoning. They generate programs with Code-LLMs and feed the programs into an external interpreter to derive the answer.~\citet{madaan2022language, wang2022code4struct} leverage the text-to-structure translation ability of Code-LLMs to perform structural prediction tasks. They ask models to generate structures in the form of code, and convert the generated code into the task output format. In addition,~\citet{Hu2022InContextLF} takes advantages of Code-LLMs on text-to-SQL generation. 
Different from them, we leverage the causal reasoning ability of Code-LLMs, and ask them to generate natural language events given the causal structure.
\subsection{Prompts}
\begin{figure*}[ht]
    \centering
    \includegraphics[width=\textwidth]{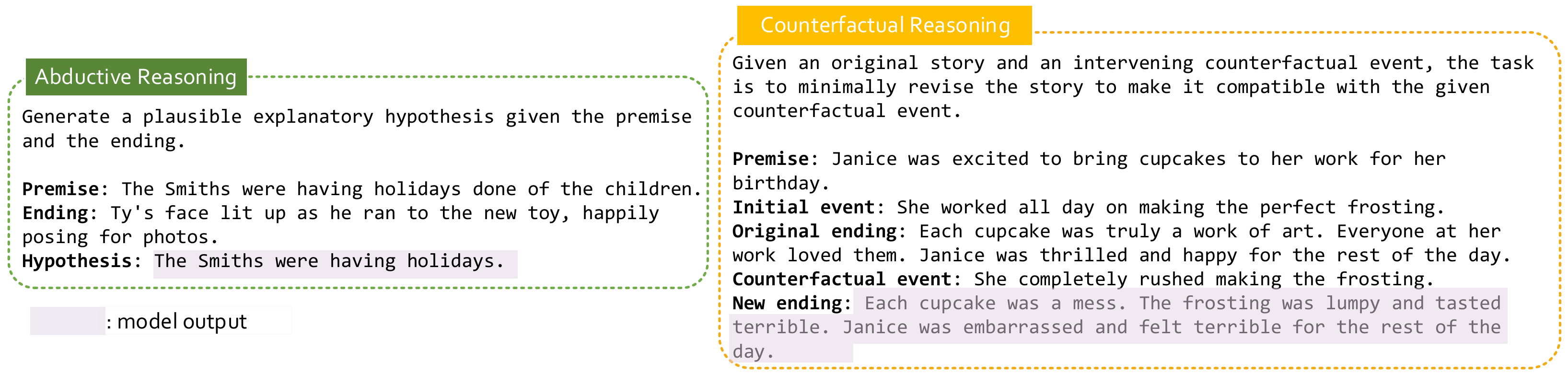}
    \caption{Example text prompts of abductive reasoning and counterfactual reasoning.} 
    \label{fig-text}
\end{figure*}
\begin{figure*}[th]
    \centering
    \includegraphics[width=0.9\textwidth]{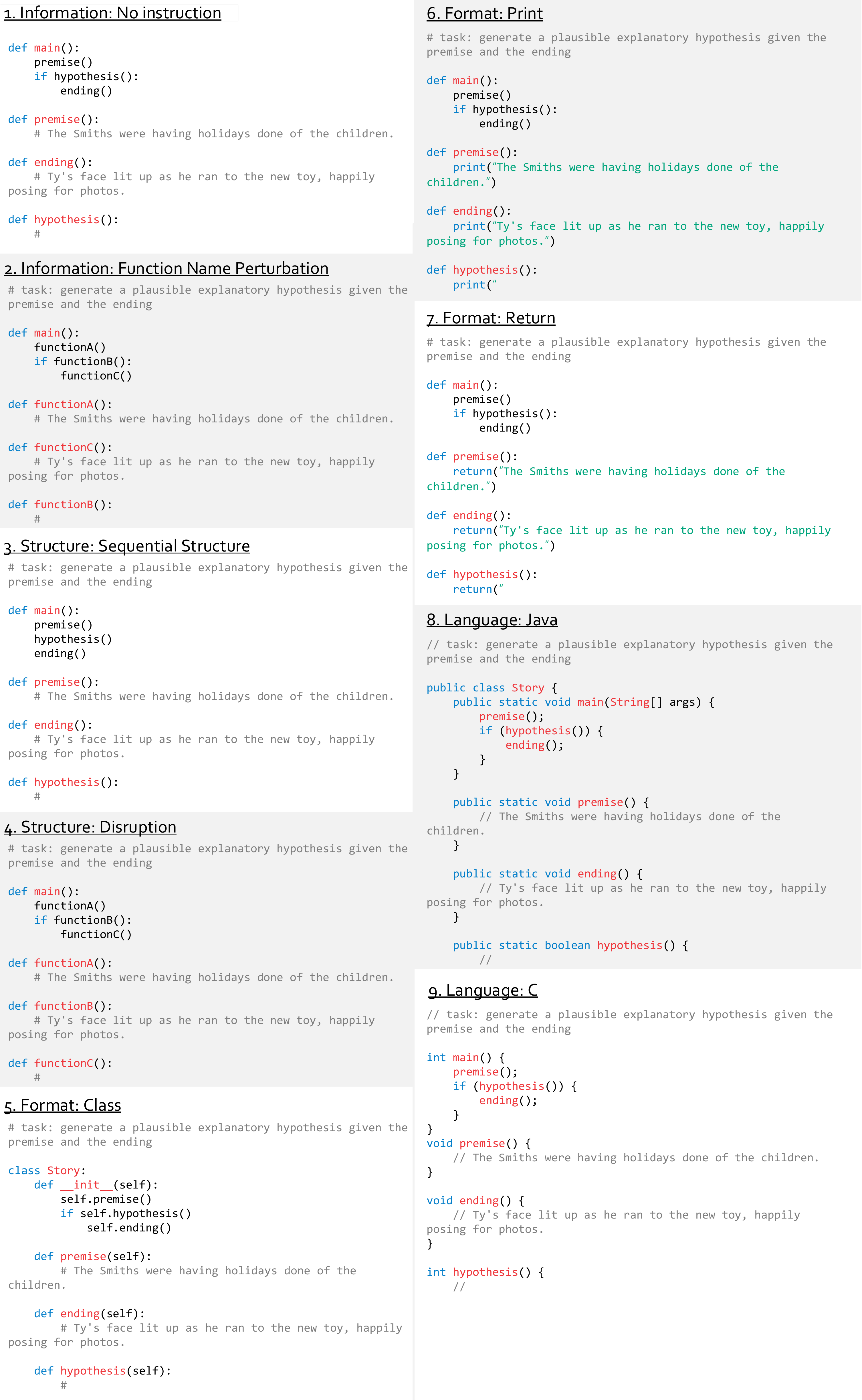}
    \caption{Examples of code prompt interventions in abductive reasoning.} 
    \label{fig-aspect}
\end{figure*}
Figure~\ref{fig-text} demonstrates the prompts of probing \textsc{Davinci}. Specifically, the language conversion is made automatically by \textsc{Codex} with the instruction \texttt{\# python to java/c}. Figure~\ref{fig-aspect} shows the interventions on code prompts for abductive reasoning.

\subsection{Models for Comparison}
\label{sec-appendix-model}
We compare with previous unsupervised methods on the two tasks, including \textsc{Delorean}~\cite{qin2020back}, \textsc{Cold}~\cite{qin2022cold}, and \textsc{Diffusion}~\cite{li2022diffusion} on abductive reasoning; and CGMH~\cite{miao2019cgmh},  \textsc{EduCat}~\cite{chen2022unsupervised}, \textsc{Delorean}, and \textsc{Cold} on counterfactual reasoning.
Among them, \textsc{Delorean} and \textsc{Cold} are constraint-based models. They regard the task requirements as constraints (for example, the generated text should be consistent with the premise, and coherent with the ending in the abductive reasoning task), and iteratively update text representation to meet the constraints. CGMH and \textsc{EduCat} are editing-based models targeted for counterfactual reasoning. They start from the original ending and edit it to meet the counterfactual context. \textsc{Diffusion} builds a controllable LM based on continuous diffusions to perform control tasks including abductive reasoning. 

\subsection{Additional Results}
\begin{table}[th]
    \centering
    \small
    \scalebox{0.9}{
    \begin{tabular}{lcc}
    \toprule
    & \textbf{Min-Edit} & \textbf{BERTScore} \\
    \midrule
    \textsc{Delorean} & 52.9 & 73.7 \\
    \textsc{Cold} & 56.8 & 73.5 \\
    \textsc{Codex} & \textbf{58.0} & \textbf{79.5} \\
    \bottomrule
    \end{tabular}
    }
    \caption{Counterfactual reasoning results in the first-sentence setting (\%).}
    \label{table-1sent}
\end{table}
\begin{table*}[th]
    \centering
    \small
    \scalebox{0.9}{
    \begin{tabular}{llccc}
    \toprule
    & & \textbf{BLEU$_4$} & \textbf{ROUGE$_\textrm{L}$} & \textbf{BERTScore} \\
    \midrule
    & \textsc{Codex} & 66.8 & 70.0 & 82.5 \\
    \midrule
    \rowcolor[gray]{0.95} &  No Instruction & 55.4 & 60.1 & 77.0  \\
    \rowcolor[gray]{0.95} \multirow{-2}{*}{\textbf{Information}} & Function Name Perturbation & 65.4 & 69.0 & 82.2 \\
    & Sequential Structure & 43.4 & 50.2 & 68.2 \\
    \multirow{-2}{*}{\textbf{Structure}} & Disruption & 16.0 & 23.5 & 55.2 \\
    \rowcolor[gray]{0.95} & Class & 63.6 & 67.4 & 81.1 \\
    \rowcolor[gray]{0.95}  & Print & \textbf{73.3} & \textbf{74.7} & \textbf{85.3} \\
    \rowcolor[gray]{0.95} \multirow{-3}{*}{\textbf{Format}} & Return & 69.4 & 70.5 & 83.0 \\
    & Java & 71.1 & 73.5 & 84.5 \\
    \multirow{-2}{*}{\textbf{Language}} & C & 71.9 & 74.2 & 85.0 \\
    \bottomrule
    \end{tabular}
    }
    \caption{Intervention results on counterfactual reasoning (\%).}
    \label{table-aspect-c}
\end{table*}
\begin{table*}[ht]
    \centering
    \small
    \scalebox{0.9}{
    \begin{subtable}[t]{0.57\textwidth}
        \centering
        \setlength{\tabcolsep}{3.5pt}
        \begin{tabular}{lcccc}
        \toprule
        & \textbf{BLEU$_4$} & \textbf{ROUGE$_\textrm{L}$} & \textbf{CIDEr} & \textbf{BERTScore} \\
        \midrule
        \textsc{Davinci$_{002}$} & 8.2 & 33.5 & 55.9 & 61.7 \\
        \textsc{Codex} & \textbf{17.9} & \textbf{42.3} & \textbf{91.7} & \textbf{67.1} \\
        \bottomrule
        \end{tabular}
        \caption{Abductive reasoning.}
    \end{subtable}
    \hspace{0.5em}
    \begin{subtable}[t]{0.41\textwidth}
        \centering
        \setlength{\tabcolsep}{3.1pt}
        \begin{tabular}{lccc}
        \toprule
        & \textbf{BLEU$_4$} & \textbf{ROUGE$_\textrm{L}$} & \textbf{BERTScore} \\
        \midrule
        \textsc{Davinci$_{002}$} & 53.5 & 58.8 & 76.0 \\
        \textsc{Codex} & \textbf{74.3} & \textbf{76.2} & \textbf{86.1} \\
        \bottomrule
        \end{tabular}
        \caption{Counterfactual reasoning.}
    \end{subtable}
    }
    \caption{Evaluation results in the one-shot setting (\%).}
    \label{table-1shot}
\end{table*}
\paragraph{First-Sentence Setting of Counterfactual Reasoning}
Endings in the original counterfactual reasoning data TimeTravel are of three sentences. Due to the computation constraint of \textsc{Cold}~\cite{qin2022cold}, it is evaluated in a first-sentence setting: only the first sentence of the original ending is used, and models are asked to generate a one-sentence counterfactual ending. We conduct experiments in the first-sentence setting with the metrics used in~\citet{qin2022cold}. As shown in Table~\ref{table-1sent}, \textsc{Codex} outperforms previous methods in this setting.

\paragraph{Intervention on Counterfactual Reasoning}
Table~\ref{table-aspect-c} demonstrates the intervention results on counterfactual reasoning. The observations are similar to those in the abductive reasoning task: changes in the programming structure affect \textsc{Codex}'s performance largely, changes in the information affect less, and \textsc{Codex} is robust towards format and language changes.

\paragraph{One-shot Setting}
We also conduct experiments in the one-shot setting. Models are shown with one demonstration example in the in-context learning manner, and the example is identical among the models. As shown in Table~\ref{table-1shot}, both \textsc{Davinci$_{002}$} and \textsc{Codex} are better than in the zero-shot setting, while \textsc{Codex} still largely outperforms \textsc{Davinci$_{002}$}, showing that the advantage of \textsc{Codex} is robust across different settings.

\subsection{Case Study}
\label{sec-appendix-case}
\begin{table*}[ht]
    \centering
    \small
    \renewcommand{\arraystretch}{1.2}
    \begin{tabularx}{\textwidth}{X}
    \toprule
    \textbf{Abductive Reasoning} \\
    \midrule
    \rowcolor[gray]{0.95} \textbf{Premise:} Angie went to a cocktail party hosted by her best friend.\\
    \rowcolor[gray]{0.95} \textbf{Ending:} Angie decided to be quiet about what she overheard.\\
    \textbf{Hypothesis [\textsc{Codex}]:} Angie overheard her best friend talking about her behind her back.\\
    \textbf{Hypothesis [\textsc{Davinci$_{002}$}]:} Angie decided to be quiet about what she overheard because she didn't want to start any drama with her best friend.\\
    \textbf{Hypothesis [\textsc{Davinci$_{003}$}]:} Angie overheard a conversation between two people at the party that made her uncomfortable, so she decided to stay quiet and not tell anyone what she heard.\\
    \rowcolor[gray]{0.95} \textbf{Hypothesis [\textsc{ChatGPT}]:} Angie overheard something that she felt was better left unsaid.\\
    \rowcolor[gray]{0.95} \textbf{Hypothesis [\textsc{GPT-4}]:} Angie overheard a sensitive conversation that she felt was not her place to share.\\
    \rowcolor[gray]{0.95} \textbf{Hypothesis [\textsc{Bard}]:} Angie overheard something at the party that she did not want to share with her best friend.\\
    \textbf{Hypothesis [\textsc{CodeGen}]:} Angie's best friend was a bartender.\\
    \textbf{Hypothesis [\textsc{StarCoder}]:} If judge is cheating on Annie, then judge was correct in trolling an English major. \\
    \midrule
    \textbf{Counterfactual Reasoning} \\
    \midrule
    \rowcolor[gray]{0.95} \textbf{Premise:} My grandpa took my brothers and I fishing in the summer.\\
    \rowcolor[gray]{0.95} \textbf{Initial Event:} We went to a small pond close to his house. \\
    \rowcolor[gray]{0.95} \textbf{Original Ending:} He set up our fishing poles and we began to fish. It was so boring when nobody caught any fish. We were disappointed to go home empty handed that day. \\
    \rowcolor[gray]{0.95} \textbf{Counterfactual Event:} We caught four huge trouts within the first hour. \\
    \textbf{Counterfactual Ending [\textsc{Codex}]:} He set up our fishing poles and we began to fish. It was so exciting when we caught four huge trouts within the first hour. We were happy to go home with four trouts that day. \\
    \textbf{Counterfactual Ending [\textsc{Davinci$_{002}$}]:} We were so excited to have caught four trouts within the first hour! We were proud to show our grandpa our catches. \\
    \textbf{Counterfactual Ending [\textsc{Davinci$_{003}$}]:} He set up our fishing poles and we began to fish. To our pleasant surprise, we caught four huge trouts within the first hour. We were delighted to take home our catches that day.\\
    \rowcolor[gray]{0.95} \textbf{Counterfactual Ending [\textsc{ChatGPT}]:} As soon as we cast our lines, we felt a tug and reeled in four huge trouts within the first hour. We couldn't believe our luck and spent the rest of the day catching even more fish. Grandpa was proud of us and we were thrilled to bring home a big haul for dinner. It was a memorable day of fishing that we would never forget.\\
    \rowcolor[gray]{0.95} \textbf{Counterfactual Ending [\textsc{GPT-4}]:} He set up our fishing poles and we began to fish. To our surprise, we caught four huge trouts within the first hour. We were thrilled with our success and couldn't wait to show off our catch when we got home.\\
    \rowcolor[gray]{0.95} \textbf{Counterfactual Ending [\textsc{Bard}]:} We caught four huge trouts within the first hour! We were so excited that we forgot all about being bored. We had a great time fishing with grandpa that day.\\
    \textbf{Counterfactual Ending [\textsc{CodeGen}]:} He set up our fishing poles and we began to fish. It was so boring when nobody caught any fish. We were disappointed to go home empty handed that day. We caught four huge trouts within the first hour.\\
    \textbf{Counterfactual Ending [\textsc{StarCoder}]:} We were finally happy to catch all of the trouts. And while he is a very kind man, you will never see him again. We will always love our old family in China better than ever before.\\
    \bottomrule
    \end{tabularx}
    \caption{Examples of model generations.}
    \label{table-case}
\end{table*}
We randomly select some generation examples and demonstrate them in Table~\ref{table-case}. Comparing \textsc{Codex} and \textsc{Davinci}, \textsc{Codex} generations are more coherent with the context, while \textsc{Davinci} sometimes cannot take into account the premise. \textsc{Codex} also understands the task instruction well and better preserves the original ending in counterfactual reasoning. 
Generations of more powerful LLMs like \textsc{ChatGPT} and \textsc{GPT-4} are coherent with the context, but they add much detail and barely keep the original ending. Although open source Code-LLMs like \textsc{CodeGen} and \textsc{StarCoder} can follow the code prompts and generate sentences in the required format, most of their outputs are inconsistent with the premise and the ending.

\end{document}